\documentclass[runningheads]{llncs}
\usepackage[T1]{fontenc}
\usepackage{graphicx}
\usepackage{lipsum}
\usepackage{amsmath}
\usepackage{amssymb}

\usepackage{booktabs}
\usepackage{xcolor}
\usepackage{graphicx}
\usepackage{pgfplots}
\pgfplotsset{compat=1.18}
\usepackage{hyperref}
\hypersetup{
    colorlinks=true,
    urlcolor=blue,
    linkcolor=black,
    citecolor=black
}

\begin{document}
\title{PET Quantification-Preserving Registration for Longitudinal Whole-Body PSMA PET/CT}
\titlerunning{Quantification-Preserving PSMA PET/CT Registration}
\author{Fryderyk K{\"o}gl\inst{1,2,3}\and
Anna Reithmeir\inst{1,2,3}\and
Daniel M. Lang\inst{1}\and
Emily Chan\inst{1,2}\and
Rickmer Braren\inst{4}\and
Julia A. Schnabel\inst{1,2,3,5}}

\authorrunning{F. K{\"o}gl et al.}

\institute{\mbox{\textsuperscript{1} Helmholtz Munich},
\mbox{\textsuperscript{2} Technical University of Munich},
\mbox{\textsuperscript{3} Munich Center for Machine Learning},
\mbox{\textsuperscript{4} University Medical Center Hamburg-Eppendorf},
\mbox{\textsuperscript{5} King's College London}}
\maketitle              %
\begin{abstract}
We adapt the diffeomorphic coarse-to-fine LapIRN architecture for longitudinal whole-body PSMA PET/CT registration in the Learn2Reg 2026 PSMAReg challenge.
The method combines affine pre-registration, multi-modal deformable registration, and test-time instance optimization.
We further introduce PET biomarker preservation losses for metabolic tumor volume (MTV) and total lesion glycolysis (TLG), including Jacobian-based constraints, and a per-structure rigidity regularization for skeletal structures.
The submitted configuration achieved $81.0\%$ DSC and $6.17$\,mm HD95, with MTV and TLG errors of $3.3\%$ and $3.3\%$, respectively, on our evaluation protocol.
Code is available at \url{https://github.com/koegl/psmareg}.
\end{abstract}
\section{Introduction}

Prostate-specific membrane antigen (PSMA) PET/CT has become a central imaging modality for staging, treatment planning, radiopharmaceutical therapy, and post-therapy response assessment in prostate cancer~\cite{ref_psma_clinical}.
In these settings, patients often undergo multiple scans over time, comprising a baseline scan before therapy and one or more follow-up scans.
Accurate baseline-to-follow-up registration can support the comparison of disease burden, the localization of uptake changes, the assessment of lesion progression or response, dose accumulation, and quantitative longitudinal analysis~\cite{ref_psmareg_challenge}.
For such analyses, the deformation should align anatomy without introducing artificial changes in PET-derived tumor burden measures.
The Learn2Reg~2026 PSMAReg challenge~\cite{ref_learn2reg_alessa,ref_learn2reg_hansen,ref_psmaregdata} formalizes this problem as the estimation of a dense displacement field that warps each follow-up (moving) scan to its baseline (fixed) scan.
The challenge evaluates registration accuracy using organ overlap and boundary distance, deformation regularity, and preservation of two PET-derived biomarkers: metabolic tumor volume (MTV) and total lesion glycolysis (TLG).

Deformable registration can be formulated as iterative optimization of a regularized objective functional~\cite{ref_demons,ref_syn,bspline}, which can achieve high accuracy but is computationally demanding for whole-body images.
Learning-based methods instead amortize this optimization into network weights and replace per-pair iteration with a single forward pass at inference~\cite{ref_voxelmorph_cvpr,ref_svf_voxelmorph,ref_transmorph}.
We therefore build on LapIRN~\cite{ref_lapirn}, whose coarse-to-fine Laplacian pyramid is well suited to recovering the large deformations expected between longitudinal whole-body scans while maintaining smooth and invertible transformations.

In this work, we adapt LapIRN to longitudinal whole-body PSMA PET/CT registration and extend it with PET biomarker preservation losses, local bone rigidity regularization, and test-time instance optimization.

\begin{figure}[t]
\centering
\includegraphics[width=\linewidth]{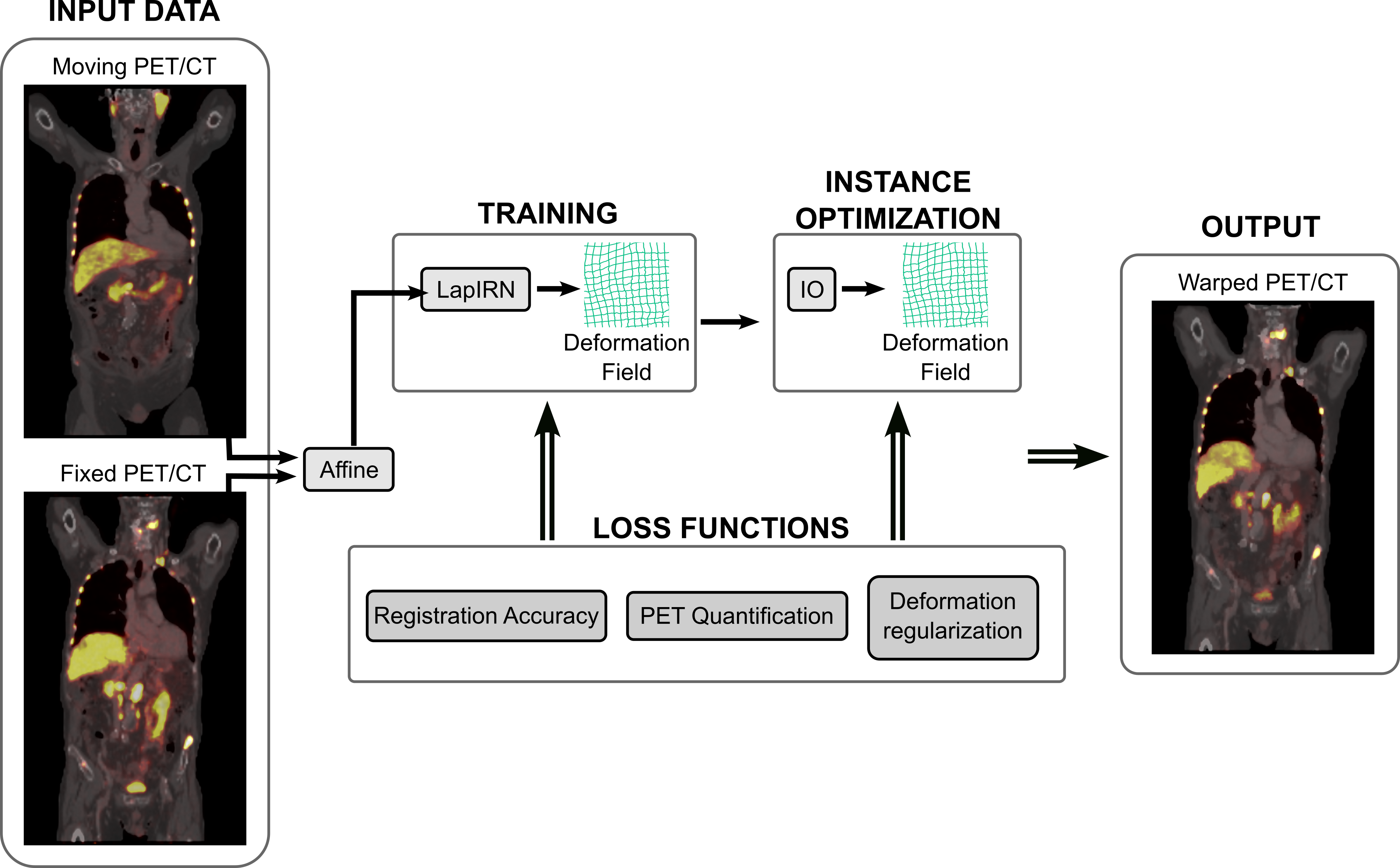}
\caption{Overview of the proposed registration pipeline.
The moving (follow-up) and fixed (baseline) PET/CT scans are first coarsely aligned by an affine pre-registration, and the resulting pair is then passed to the coarse-to-fine LapIRN network, which predicts a dense deformation field (green grid).
Finally, after the LapIRN stage, we refine this field per pair by instance optimization (IO) and resample the moving scan with it into the fixed space (warped PET/CT).
Both stages are driven by the same objective, a weighted sum of three groups of terms: registration accuracy, PET quantification, and deformation regularization (Sec.~\ref{sec:losses}).}
\label{fig:pipeline}
\end{figure}

\section{Methods}

\subsection{Data and Preprocessing}
\label{sec:data_and_preprocessing}

We used the PSMAReg dataset of the Learn2Reg 2026 challenge~\cite{ref_psmareg_challenge}, a longitudinal whole-body PSMA PET/CT cohort in which each subject provides a baseline scan and one or more follow-up scans.
Each scan contains CT and PSMA PET channels; follow-up scans were treated as moving images and baseline scans as fixed images.
The images are provided at $2.7344\times2.7344\times3.27$~mm resolution and cropped to $192\times192\times288$ voxels.
We therefore applied no further geometric preprocessing.
CT was clipped to $[-1000, 1500]$~HU and PET SUV to $[0, 20]$, followed by rescaling to $[0,1]$.

Of the 358 released training cases, we used the 146 paired cases and split them patient-wise into 117 training and 29 validation cases.
The 20 challenge validation cases, for which no labels are released, were used as our test set.
For training and validation, we used the CT organ and PET lesion labels provided by the organizers.
For the test set, we generated surrogate CT organ labels using TotalSegmentator~\cite{ref_totalsegmentator} and PET lesion labels using the winning autoPET III submission~\cite{rokuss2024fdgpsmahitchhikersguide}.
Following the ConvexAdam~\cite{ref_convexadam} baseline provided by the challenge organizers, we additionally removed the patient bed and normalized the CT images.

\subsection{Registration Pipeline}
\label{sec:pipeline}
Our pipeline consists of CT-based affine pre-registration, LapIRN-based deformable registration, and optional test-time instance optimization (Fig.~\ref{fig:pipeline}).
Throughout, $I$ denotes the moving PET image, $M$ its lesion mask, $\varphi_{\mathrm{aff}}$ the affine pre-registration, $\varphi_{\theta}$ the network-predicted deformation, and $\varphi$ their composition.
$W$ is the lesion mask warped into the fixed frame.

\subsubsection{Step 1: Affine pre-registration.}
For each pair, ANTs~\cite{ants} estimates an affine transform from the CT images alone, using body masks, CT windowing to $[-1000,1000]$~HU, half-resolution images, and Mattes mutual information.
The transform is converted to a dense displacement field and applied to the moving scan.

\subsubsection{Step 2: Deformable registration.}
The deformable stage uses LapIRN~\cite{ref_lapirn}, with three pyramid levels operating at quarter, half, and full resolution.
Each level predicts a stationary velocity field, adds it to the upsampled velocity from the coarser level, and exponentiates the result by scaling and squaring to obtain a diffeomorphic $\varphi_{\theta}$.
To handle PSMA PET/CT, each sub-network receives four channels: fixed CT and PET, and affinely aligned moving CT and PET.
The similarity loss is computed only on CT, because PET uptake may change after therapy; PET is used as input and in the preservation terms of Sec.~\ref{sec:losses}.
The levels are trained sequentially, with each finer level initialized from the previous one.

\subsubsection{Step 3: Instance optimization.}
At test time, each pair is refined by optimizing a residual stationary velocity field on top of the network prediction.
The residual is initialized to zero and exponentiated by scaling and squaring, preserving diffeomorphism while only correcting the predicted field.
Optimization uses the training objective (Sec.~\ref{sec:losses}); each iterate is scored by DSC, HD95, MTV, and TLG, and the best-scoring iterate is retained.

\subsection{Training Loss Function for deformable stage}
\label{sec:losses}
We use a weighted sum of three loss groups, one per evaluation criterion:
\begin{equation}
  \begin{aligned}
    \mathcal{L} = \;
      & \underbrace{\lambda_{\mathrm{NCC}}\,\mathcal{L}_{\mathrm{NCC}}
      + \lambda_{\mathrm{DSC}}\,\mathcal{L}_{\mathrm{DSC}}}_{\text{registration accuracy}} \\
      & + \underbrace{\lambda_{\mathrm{MTV}}\,\mathcal{L}_{\mathrm{MTV}}
      + \lambda_{\mathrm{MTV\text{-}J}}\,\mathcal{L}_{\mathrm{MTV\text{-}J}}
      + \lambda_{\mathrm{jac}}\,\mathcal{L}_{\mathrm{jac}}
      + \lambda_{\mathrm{TLG}}\,\mathcal{L}_{\mathrm{TLG}}}_{\text{PET quantification}} \\
      & + \underbrace{\lambda_{\mathrm{rigid}}\,\mathcal{L}_{\mathrm{rigid}}
      + \lambda_{\mathrm{smooth}}\,\mathcal{L}_{\mathrm{smooth}}
      + \lambda_{\mathrm{NDV}}\,\mathcal{L}_{\mathrm{NDV}}}_{\text{deformation regularization}} .
  \end{aligned}
  \label{eq:total}
\end{equation}
Each group is described below.
All terms are evaluated on the composed transformation $\varphi$.
Only the full-resolution level is trained with the complete objective.
The two coarser levels (quarter and half resolution) use only the similarity, label, and regularization terms; thus $\lambda_{\mathrm{MTV}} = \lambda_{\mathrm{MTV\text{-}J}} = \lambda_{\mathrm{jac}} = \lambda_{\mathrm{TLG}} = \lambda_{\mathrm{rigid}} = 0$.
We omit the PET and bone-specific terms at these resolutions because downsampled lesions and thin ribs span only a few voxels, so volume ratios and per-structure rigid fits mainly reflect interpolation rather than anatomy.

\subsubsection{Registration Accuracy.}
Image similarity is measured using local normalized cross-correlation (NCC), computed as a multi-resolution sum over as many scales as the respective pyramid level has resolutions, i.e.\ one scale at the coarsest and three at the full-resolution level.
NCC is computed on the CT channel only, since PET uptake can change substantially between time points, and a PET similarity term could therefore encourage matching disease-related uptake rather than anatomy.
We additionally use a DSC loss for weak supervision from the CT organ labels.

\subsubsection{Preservation of PET quantification.}

The challenge evaluates the net change in metabolic tumor volume (MTV) and total lesion glycolysis (TLG) under the estimated transformation.
All terms are evaluated on the composed transformation $\varphi$ and the original moving lesion mask $M$.

To additionally constrain the deformation within the lesion, we use $\mathcal{L}_{\mathrm{MTV\text{-}J}}$, which penalizes deviations of the mean Jacobian determinant over the warped lesion region $W$ from one.
A third term, $\mathcal{L}_{\mathrm{jac}}$, penalizes per-voxel deviations of the Jacobian determinant from one and therefore discourages local compensating compression and expansion.

TLG is penalized analogously by $\mathcal{L}_{\mathrm{TLG}}$, the relative change in PET intensity mass within the lesion before and after warping.
Because both the PET image and the lesion mask are transformed, this term captures changes in lesion volume and in intensity interpolation.
All four PET preservation terms are used during training and instance optimization.

\subsubsection{Bone rigidity.}
Many lesions lie within skeletal structures, where rigid motion is physiologically plausible and preserves both volume and intensity ($\det J_{\varphi} = 1$).
We therefore constrain each skeletal structure to undergo rigid motion.
For the 61 skeletal TotalSegmentator labels, we fit the displacement of each structure independently with a rigid transformation using closed-form Kabsch alignment~\cite{ref_kabsch,ref_umeyama} and penalize the residual.
This constrains the deformation within each structure while allowing independent motion between structures.

\subsubsection{Deformation regularization.}
We regularize the displacement field using a diffusion term, $\mathcal{L}_{\mathrm{smooth}}$, defined as the mean squared spatial gradient of the displacement, together with a penalty for non-diffeomorphic voxels.
For the latter, we differentiably reproduce the challenge's non-diffeomorphic volume (NDV) metric, which decomposes each voxel into six tetrahedra and accumulates the negative part of their Jacobian determinants over the body mask~\cite{Liu2024}.
The resulting $\mathcal{L}_{\mathrm{NDV}}$ directly corresponds to the scored percentage of non-diffeomorphic volume and provides gradients at the locations where folding occurs.

\subsection{Submission container}
\label{sec:container}
The submitted container must produce all labels required by instance optimization and register each pair within the 90~s time limit.
It therefore differs from the validation configuration only in this final stage.
PET lesion masks are predicted by a nnU-Net~\cite{nnunet} trained on the challenge dataset with progressive patch-size growing~\cite{ref_progressive_patch}.
CT organ labels are generated with TotalSegmentator in fast mode.
Instance optimization runs until the time limit is reached.
The affine and network-predicted deformable fields are unchanged.
We report both configurations separately in Sec.~\ref{sec:evaluation}, since only the validation setting is unconstrained by runtime.

\subsection{Model Selection}
\label{sec:model_selection}
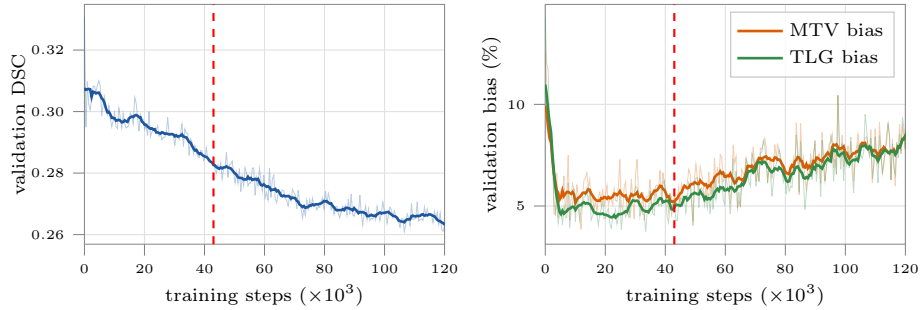
\begin{figure}[t]
\centering
\definecolor{colordsc}{rgb}{0.12,0.34,0.62}
\definecolor{colormtv}{rgb}{0.85,0.37,0.01}
\definecolor{colortlg}{rgb}{0.20,0.55,0.28}
\tikzset{selectionline/.style={red, dashed, line width=0.8pt}}
\pgfplotsset{every axis/.append style={font=\scriptsize, label style={font=\scriptsize}, tick label style={font=\tiny}}}
\begin{tikzpicture}
  \begin{axis}[
    width=0.39\linewidth,
    height=0.26\linewidth,
    scale only axis,
    xlabel={training steps ($\times 10^{3}$)},
    ylabel={validation DSC},
    ylabel near ticks,
    ylabel shift=-4pt,
    xlabel shift=-2pt,
    ymin=0.256880,
    ymax=0.334841,
    xmin=0,
    xmax=120,
    xtick distance=20,
    yticklabel style={/pgf/number format/fixed, /pgf/number format/precision=2, /pgf/number format/zerofill},
    tick align=outside,
    tick pos=left,
    grid=major,
    grid style={draw=black!12},
    axis line style={draw=black!45},
  ]
    \addplot [draw=colordsc, line width=0.35pt, opacity=0.30, forget plot] table [x=step, y=dsc, col sep=comma] {figures/dsc_plateau_data.csv};
    \addplot [draw=colordsc, line width=1.0pt] table [x=step, y=dsc_smooth, col sep=comma] {figures/dsc_plateau_data.csv};
    \addplot [selectionline, forget plot] coordinates {(43.0000,0.256880) (43.0000,0.334841)};
  \end{axis}
\end{tikzpicture}%
\hfill%
\begin{tikzpicture}
  \begin{axis}[
    width=0.39\linewidth,
    height=0.26\linewidth,
    scale only axis,
    xlabel={training steps ($\times 10^{3}$)},
    ylabel={validation bias (\%)},
    ylabel near ticks,
    ylabel shift=-4pt,
    xlabel shift=-2pt,
    ymin=3.121323,
    ymax=14.915233,
    xmin=0,
    xmax=120,
    xtick distance=20,
    yticklabel style={/pgf/number format/fixed, /pgf/number format/precision=0, /pgf/number format/zerofill},
    tick align=outside,
    tick pos=left,
    grid=major,
    grid style={draw=black!12},
    axis line style={draw=black!45},
    legend pos=north east,
    legend cell align=left,
    legend style={draw=black!25, fill=white, fill opacity=0.85, text opacity=1, font=\scriptsize, inner sep=2pt},
  ]
    \addplot [draw=colormtv, line width=0.35pt, opacity=0.30, forget plot] table [x=step, y=mtv, col sep=comma] {figures/dsc_plateau_data.csv};
    \addplot [draw=colormtv, line width=1.0pt] table [x=step, y=mtv_smooth, col sep=comma] {figures/dsc_plateau_data.csv};
    \addlegendentry{MTV bias}
    \addplot [draw=colortlg, line width=0.35pt, opacity=0.30, forget plot] table [x=step, y=tlg, col sep=comma] {figures/dsc_plateau_data.csv};
    \addplot [draw=colortlg, line width=1.0pt] table [x=step, y=tlg_smooth, col sep=comma] {figures/dsc_plateau_data.csv};
    \addlegendentry{TLG bias}
    \addplot [selectionline, forget plot] coordinates {(43.0000,3.121323) (43.0000,14.915233)};
  \end{axis}
\end{tikzpicture}
\caption{Registration accuracy and biomarker preservation peak at different checkpoints. Validation CT Dice (left) continues to improve after MTV and TLG bias (right) reach their minima. The red dashed line indicates the checkpoint selected using the challenge composite ranking score.}
\label{fig:selection}
\end{figure}

We use the challenge composite ranking score for model selection, which combines registration accuracy, PET biomarker preservation, and deformation regularity with weights of $0.4$, $0.4$, and $0.2$, respectively.
This score is used for both checkpoint and final variant selection.
As shown in Fig.~\ref{fig:selection}, validation DSC continues to improve after MTV and TLG errors reach their minima, making registration accuracy alone unsuitable for model selection.

\subsection{Baselines}
\label{sec:baselines}
We compare against two references and two competing methods, all evaluated on the same pairs and under the same protocol as in Sec.~\ref{sec:evaluation}.

As a classical iterative baseline, we ran NiftyReg~\cite{MODAT2010,modat2014}, using \texttt{reg\_aladin} for the affine stage followed by \texttt{reg\_f3d} initialized with it, both with default parameters.
As a learning-free optimization baseline we ran ConvexAdam, in the form provided by the challenge organizers as the official baseline of the task.

\subsection{Implementation Details}
\label{sec:infrastructure}
Augmentations were chosen to mimic longitudinal variability: left--right flipping ($p=0.5$), CT intensity shifting by $\pm0.02$ in the normalized window (about $\pm50$~HU), CT scaling from $0.9$ to $1.1$, and PET scaling from $0.85$ to $1.15$ without additive shifts.
To simulate a field-of-view mismatch, we crop up to 40 axial slices from either end of both scans, followed by independent moving and fixed crops of up to 10 additional slices.

The three pyramid levels are trained sequentially with Adam for $60$k, $60$k and $120$k steps, using learning rates of $3\cdot10^{-4}$, $2\cdot10^{-4}$ and $2.5\cdot10^{-4}$, batch size one, and gradient accumulation over four steps.
Each level uses a five-epoch linear warmup; the preceding level is frozen for the first ten epochs and then fine-tuned jointly.
The convolutional trunk runs in \texttt{bfloat16}, while transformations and losses remain in single precision to avoid numerical errors in scaling and squaring and Jacobian computations.
The loss weights were chosen by the sweep of Sec.~\ref{sec:evaluation}.

All models were trained and evaluated on a single NVIDIA H100 80\,GB HBM3 GPU (driver 610.57.04) using Python~3.11, PyTorch~2.6.0~\cite{ref_pytorch}, CUDA~12.4, ANTsPy~0.6.3~\cite{ants}, NiBabel~5.4.2, NumPy~2.4.4 and SciPy~1.17.1.
Training all three levels took 2d 6h 7m 6s.

\section{Results}

\subsection{Evaluation Protocol}
\label{sec:evaluation}

We report three configurations of our method.
\emph{Ours (container)} is the submitted method of Sec.~\ref{sec:container}; it is the configuration evaluated against the baselines because it matches the hidden-test submission.
\emph{Ours (validation)} uses the unconstrained pipeline of Sec.~\ref{sec:pipeline}, while \emph{ours (no IO)} removes instance optimization to isolate its contribution.
Significance markers are shown only for comparisons between \emph{ours (container)} and the baselines.

Table~\ref{tab:baselines} is evaluated on the 20 official challenge validation cases using our surrogate labels and local evaluator.
These results are therefore separate from the official Codabench validation leaderboard, which uses the organizers' scoring setup.

\subsection{Comparison against Baselines}

We swept the loss weights and the instance-optimization settings and selected the one reported below with the challenge ranking of Sec.~\ref{sec:model_selection}, applied to the surrogate protocol above.
Table~\ref{tab:baselines} places the three configurations against the baselines.
The container configuration improves DSC from $73.3\%$ (NiftyReg, the strongest baseline) to $81.0\%$ and reduces HD95 from $7.09$ to $6.17$\,mm, while also reducing the MTV and TLG errors and the non-diffeomorphic volume from $1525$ to $864$\,ppm.

\begin{table}[t]
\centering
\setlength{\belowcaptionskip}{5pt}
\caption{Our method against the baselines on the official challenge validation set ($n=20$ cases), evaluated against the surrogate labels of Sec.~\ref{sec:evaluation}. Every cell is a mean over the cases, runtime included: DSC and the MTV/TLG errors in \%, NDV in ppm, and time in seconds per image pair on the hardware of Sec.~\ref{sec:infrastructure}. Bold marks the best value in each column. $^{*}$ marks metrics on which \emph{ours (container)} is significantly better than the baseline in that row (paired Wilcoxon signed-rank, Holm-corrected within each metric, $p<0.05$); runtime is not tested. Dashes mark metrics undefined for a baseline.}
\label{tab:baselines}
\small
\setlength{\tabcolsep}{4pt}
\begin{tabular}{lcccccc}
\toprule
Method & \rotatebox[origin=l]{60}{\shortstack{DSC\\(\%) $\uparrow$}} & \rotatebox[origin=l]{60}{\shortstack{HD95\\(mm) $\downarrow$}} & \rotatebox[origin=l]{60}{\shortstack{MTV error\\(\%) $\downarrow$}} & \rotatebox[origin=l]{60}{\shortstack{TLG error\\(\%) $\downarrow$}} & \rotatebox[origin=l]{60}{\shortstack{NDV\\(ppm) $\downarrow$}} & \rotatebox[origin=l]{60}{\shortstack{Time\\(s) $\downarrow$}} \\
\midrule
Initial & 14.8$^{*}$ & 30.9$^{*}$ & -- & -- & -- & -- \\
Affine & 52.9$^{*}$ & 11.7$^{*}$ & 5.47 & 4.70 & -- & \textbf{19.1 $\pm$ 0.920} \\
ConvexAdam & 63.0$^{*}$ & 8.41$^{*}$ & 5.10 & 4.15 & 100 & 22.8 $\pm$ 1.06 \\
NiftyReg & 73.3$^{*}$ & 7.09$^{*}$ & 4.46 & 8.68$^{*}$ & 1525$^{*}$ & 49.7 $\pm$ 0.480 \\
\midrule
Ours (no IO) & 76.5 & 6.65 & 3.22 & 5.08 & \textbf{0.00} & 19.5 $\pm$ 1.02 \\
Ours (container) & 81.0 & 6.17 & 3.29 & 3.31 & 864 & 89.8 $\pm$ 2.69 \\
Ours (validation) & \textbf{88.3} & \textbf{4.80} & \textbf{1.37} & \textbf{2.13} & 70.9 & 383 $\pm$ 7.77 \\
\bottomrule
\end{tabular}
\end{table}

\subsection{Leaderboard Performance}
These leaderboard numbers are reported separately because they were computed by the official evaluation server rather than by our surrogate-label protocol.
On the validation leaderboard, our method (\emph{Ours (validation)}) achieved a Dice score of $0.789 \pm 0.0381$, HD95 of $5.99 \pm 2.39$\,mm, MTV error of $1.56 \pm 1.22\%$, TLG error of $4.18 \pm 2.49\%$, and NDV of $0.0041 \pm 0.0024\%$.

\section{Discussion}
This work adapts a diffeomorphic coarse-to-fine registration pipeline to longitudinal whole-body PSMA PET/CT, with an explicit focus on preserving PET-derived biomarkers.
Although checkpoint selection shows that registration accuracy alone is not sufficient for model selection, the final submitted configuration improves both alignment and PET preservation compared with the baselines.
Relative to the strongest baseline, it improves DSC and HD95 while reducing MTV and TLG bias, as well as non-diffeomorphic volume.
The biomarker terms are applied consistently during training and instance optimization, while the per-structure rigid fit constrains skeletal structures without relying on finite-difference stencils that may extend into surrounding soft tissue.

Several limitations qualify these results.
Evaluation relies on automatically generated surrogate labels, and the loss weights were selected on the same 20 cases used for comparison.
The reported results should therefore be interpreted as exploratory rather than confirmatory.
Hidden test-set results were unavailable at the time of writing.
The rigidity term also depends on the quality of the skeletal segmentations and is primarily intended for structures that are thin relative to the image resolution.

Finally, the 90s runtime constraint reduces the accuracy of the submitted container configuration, largely due to the segmentations required for instance optimization.
Future work will therefore focus on faster, more accurate segmentation and on evaluating biomarker preservation per lesion rather than only at the aggregate MTV and TLG levels.

\begin{credits}
% \subsubsection{\ackname}

\subsubsection{\discintname}
The authors have no competing interests to declare that are relevant to the content of this article.

\end{credits}
\bibliographystyle{splncs04-etal}

\end{document}